\documentclass[11pt]{article}

\usepackage[utf8]{inputenc}
\usepackage[semibold]{sourcesanspro}
\usepackage[default]{sourceserifpro}
\usepackage[T1]{fontenc}
\usepackage[amsthm,stix2,vvarbb]{newtxmath}
\usepackage{bm}
\usepackage{nicefrac}
\usepackage{microtype}

\usepackage[dvipsnames]{xcolor} 
\usepackage{hyperref}
\usepackage{float}
\hypersetup{
	colorlinks=true,
    linkcolor=Maroon, 
    filecolor=Maroon,
	urlcolor=BlueViolet, 
    citecolor=BlueViolet,
}
\usepackage{url, epsfig}
\usepackage{booktabs}
\usepackage{subcaption}
\usepackage{threeparttable}
\usepackage{layouts}
\usepackage[export]{adjustbox}[2011/08/13]

\usepackage[authordate-trad, natbib=true, doi=false,
isbn=false,url=false,eprint=false, backend=biber]{biblatex-chicago}
\DeclareBibliographyAlias{software}{misc}
\usepackage[parfill]{parskip}
\usepackage{graphicx}
\usepackage{relsize}
\usepackage{fancyvrb}

\usepackage[a4paper,
    left=1.25in,
    right=1.25in,
    top=0.75in,
    bottom=0.75in,
    includehead,
    includefoot
]{geometry}

\usepackage{titlesec}
\titleformat*{\section}{\large\bfseries\sffamily}
\titleformat*{\subsection}{\sffamily}
\titleformat*{\subsubsection}{\itshape\sffamily}
\titleformat*{\paragraph}{\bfseries\sffamily}
\titlelabel{\thetitle.\quad}

\usepackage{caption}
\DeclareCaptionFormat{custom}
{%
    {\small\bfseries\sffamily#1#2} {\small #3}
}
\DeclareCaptionLabelSeparator{custom}{.}
\usepackage[runin]{abstract}

\abslabeldelim{.\,\,}
\newenvironment{keywords}
{
  \makeatletter

  \abslabeldelim{:\,\,}
  \makeatother
  \abstract
}{
  \endabstract\par
}

\newcommand{\pkg}[1]{{\fontseries{m}\fontseries{b}\selectfont #1}}

\usepackage{authblk}
\newlength\aftertitskip   
\makeatletter
\def\@maketitle{
  \vspace*{-0.5in}
  \begin{flushleft}
    {\Large\bfseries\sffamily\@title\par} %
    \vskip\aftertitskip
    {\large\@author\par} %
  \end{flushleft}
}
\makeatother

\title{An Image is Worth $K$ Topics: A Visual Structural Topic Model with
Pretrained Image Embeddings}

\author[1,*]{Matías Piqueras}
\author[2]{Alexandra Segerberg}
\author[1]{Matteo Magnani}
\author[3]{Måns Magnusson}
\author[4]{Nataša Sladoje}

\affil[1]{InfoLab, Department of Information Technology, Uppsala University, Sweden}
\affil[2]{Department of Government, Uppsala University, Sweden}
\affil[3]{Department of Statistics, Uppsala University, Sweden}
\affil[4]{Centre for Image Analysis, Department of Information Technology, Uppsala University, Sweden}
\affil[*]{Corresponding author: \url{matias.piqueras@it.uu.se}}
\date{}

\begin{document}

\maketitle
\thispagestyle{empty}

\begin{abstract}
\small Political scientists are increasingly interested in analyzing visual
content at scale. However, the existing computational toolbox is still in need
of methods and models attuned to the specific challenges and goals of social and
political inquiry. In this article, we introduce a visual Structural Topic Model
(\pkg{vSTM}) that combines pretrained image embeddings with a structural topic
model. This has important advantages compared to existing approaches. First,
pretrained embeddings allow the model to capture the semantic complexity of
images relevant to political contexts. Second, the structural topic model
provides the ability to analyze how topics and covariates are related, while
maintaining a nuanced representation of images as a mixture of multiple topics.
In our empirical application, we show that the \pkg{vSTM} is able to identify
topics that are interpretable, coherent, and substantively relevant to the study
of online political communication.
\end{abstract}


\section{Introduction}\label{sec:introduction}

It is important to develop a robust capacity to analyze visual data at large
scale in the study of politics. Visual content is already known to be potent in
conveying political ideas and mobilizing support, and the deluge of images and
videos in the contemporary media environment, if anything, amplifies its role.
Since images may convey something different from text
\citep{colemanFramingPicturesOur2009}, it can be crucial to analyze visual data
directly. However, compared with the text-as-data literature that has developed
a range of specialized computational methods tailored to social scientific
purposes, the visual equivalent is still underdeveloped
\citep{williamsImagesDataSocial2020a, bucyEditorsIntroductionVisual2021,
torresLearningSeeConvolutional2022, magnaniConditionsIntegratingDeep2021,
pengAutomatedVisualAnalysis2024}.

This paper addresses the task of automatically categorizing unlabeled image data
in the presence of other analytically significant variables. The ability to
explore and analyze the distribution of visual themes in a dataset when the
themes are unknown is central to a number of analytical scenarios. Several pivot
on the relationship between visual content and other covariates, as seen for
example in studies of how various actors use visual strategies to shape public
agendas and engage audiences with their cause
\citep[e.g.,][]{kimSeeingBlackLives2023, jingSelfSupervisedVisualFeature2021a,
torresFrameworkUnsupervisedSemiSupervised2024,
laaksonenAffectiveVisualRhetoric2022, pengWhatMakesPoliticians2021,
farkasImagesPoliticiansSocial2021, moosederSocialMediaLogics2023,
qianConvergenceDivergenceCrossplatform2024}, or research that analyses the key
drivers in processes of discursive politicization or protest mobilization
\citep{knupferPoliticizationRightwingNormalization2023, oakHowRhetoricWomen2024,
luMobilizingPowerVisual2024}.

Dealing with large-scale visual data presents a distinct methodological frontier
in computational social science. Images in social and political contexts
typically entail dense semantic complexity: a basic layer with information about
elements such as color and brightness; a literal layer in which the denotative
meaning of objects comes into play; and a conceptual layer that engages the
implied relationship between objects, the scene, and the connotative meaning in
ways that invoke a semantic and cultural gap between what can be extracted and
how the image is understood in context
\citep{smeuldersContentbasedImageRetrieval2000,
vannoordSurveyComputationalMethods2022}. In this regard, political image data
presents a distinct challenge from words in text-as-data and visual features
such as color and texture in, for example, medical image analysis. Intriguingly,
however, recent developments in computer vision offer potential to broach this
frontier.

We propose a \emph{visual} Structural Topic Model (\pkg{vSTM}) that is designed
to deal with the semantic complexity of images in social and political settings,
while enabling mixed membership modeling and the capacity to capture the
relationship between a visual topic and other variables of interest. The model
is designed for high ease and low expense in use, works under conditions of
limited data access, and is flexible in that it can build on any pretrained
vision model that encodes relevant image features as a continuous vector.

The \pkg{vSTM} integrates two compelling lines of research while addressing
their respective limitations. On the one hand, it builds on state-of-the-art
image representation techniques using pretrained image embeddings to capture
semantic complexity. On the other, it leverages the literature on probabilistic
topic models, allowing images to belong to multiple topics and adding the
ability to incorporate substantive structure to connect the visual topics with
other variables of interest \citep{quinnHowAnalyzePolitical2010,
grimmerBayesianHierarchicalTopic2010, robertsStructuralTopicModels2014,
eshimaKeywordAssistedTopicModels2024}.

Combining the two trajectories faces a challenge: traditional topic models
operate on discrete features (word counts). The \pkg{vSTM} therefore builds on
an alternative formulation of the statistical underpinnings of the topic model
that accepts real-valued inputs. Specifically, we extend a mixed-membership
model that enables us to think of image embeddings as a weighted sum of topics
and topic proportions \citep{cutlerArchetypalAnalysis1994,
hellerStatisticalModelsPartial2008, giesenMixedMembershipGaussians2023}. Our
contribution on this point is two-fold. First, we introduce a hierarchical prior
for the topic proportions that links covariates to the prevalence of topics
through a generalized linear model. This is analogous to how the structural
topic model for text extends Latent Dirichlet Allocation and the Correlated
Topic Model \citep{blei2003latent, blei2006correlated}. Second, to make the
model scalable to large image datasets, we implement and demonstrate how it can
be estimated using variational and amortized inference
\citep{kingmaAutoencodingVariationalBayes2013a,
hoffmanStochasticVariationalInference2013,
kucukelbirAutomaticDifferentiationVariational2017}. This makes it feasible to
explore several topic models in large image datasets, while also facilitating
extensions and modifications of the model.

\section{From Words and Pixels to Pretrained Image Embeddings}\label{sec:embeddings}

Visual topic modeling has been explored via several tracks. In this section, we
describe some of the conceptual and practical challenges involved in
representing images numerically for topic modeling purposes, while examining
existing approaches and relevant work in the field of computer vision. We
propose integrating Deep Learning image representation techniques that
constitute state of the art in capturing semantic richness in images.
Specifically, we suggest leveraging \emph{pretrained image
embeddings}---representations extracted from pretrained vision models---as an
effective foundation for visual topic modeling in political science research.

A central question in any analysis involving visual content is how to
numerically represent images in a way that retains meaning and lends itself to
statistical analysis. Unlike textual data, where units are organized in a
natural hierarchy of meaning (e.g., letters form words and words form
sentences), digital images store information as pixels without explicitly
encoding semantic structure. This makes computational analysis challenging since
individual pixels carry almost no information about the content or meaning of an
image when viewed individually---it is from their arrangement that we can make
sense of things like textures, shapes, prominent colors, composition, and
high-level semantic categories like ``person'' or ``car''. Much work in computer
vision revolves around finding ways to turn pixels into meaningful
representations, evolving from hand-crafted features like Local Binary Patterns
\citep{ojalaComparativeStudyTexture1996} and Histogram of Gradients
\citep{dalalHistogramsOrientedGradients2005} to learned representations from
convolutional neural networks (CNN;
\citealp{lecunBackpropagationAppliedHandwritten1989,
krizhevskyImageNetClassificationDeep2012}, see also
\citealp{torresLearningSeeConvolutional2022} for a political science
introduction) and, more recently, vision transformers
\citep[ViT;][]{dosovitskiyImageWorth16x162020}. The latter approaches use data
to inform what are relevant sources of variance both in terms of low-level
patterns and high-level semantic concepts
\citep{bengioRepresentationLearningReview2013}.

To date, most attempts at visual topic modeling for political analysis sidestep
the issue by relying on text descriptions or transcripts. The few pioneers who
do tackle image data draw on the logic of text topic models.  Just as text topic
models like Latent Dirichlet Allocation \citep{blei2003latent} or the Structural
Topic Model \citep{robertsStructuralTopicModels2014} operate on a document-term
matrix that contains the frequency of words in each document, these pioneering
visual approaches count the frequency of discrete features in the images. An
important contribution is \citet{torresFrameworkUnsupervisedSemiSupervised2024},
which combines the Bag of Visual Words (BoVW) with topic modeling \citep[see
also][]{fei-feiBayesianHierarchicalModel2005,
sudderthLearningHierarchicalModels2005}. The BoVW consists of several steps,
where local features are first extracted from images, then clustered into visual
words and finally counted to record the frequency in each image
\citep{sivicVideoGoogleText2003, csurkaVisualCategorizationBags2004}. A similar
logic of counting discrete features can be seen in approaches focusing on
objects in the picture. This is exemplified by \citet{kimSeeingBlackLives2023},
who use a commercial object detection model to extract and count objects to
input into a topic model. These approaches integrate smoothly with traditional
topic models.

However, focusing on discrete features in the case of visual data risks eliding
analytically significant semantic complexity. On a basic level, images typically
contain a combination of possibly interacting continuous and discrete features.
For example, the presence of an object is a discrete event, however, its
perceived salience in the picture may depend on its size and position relative
to other objects. The BoVW used by
\citet{torresFrameworkUnsupervisedSemiSupervised2024}, relies on low-level
descriptors like Scale Invariant Feature Transforms (SIFT) that detect
distinctive patches based on edge patterns and corners
\citep{loweDistinctiveImageFeatures2004}. This identifies textural elements and
simple shapes, but typically struggles to capture higher-level semantic concepts
and ignores spatial relationships. Consequently, scenes that are fundamentally
different but share local elements (e.g., a protest, parade, and concert) may be
treated identically. The object detection approach, meanwhile, reduces an image
to counts of objects, overlooking their arrangement and composition. The
categories are easily interpretable, but limited to the objects the model is
trained to detect. It is rarely feasible to train for all objects of possible
interest, so researchers typically resort to models trained on already annotated
datasets, such as COCO  \citep{linMicrosoftCOCOCommon2014}. Such delimited
``vocabularies'' may suffice in supervised settings
\citep[see][]{scholzImprovingComputerVision2024}, but are less appropriate for
unsupervised tasks like topic modeling, since knowing which objects will be
relevant for yet-to-be-discovered topics is difficult, if not logically
impossible. Models may exclude key categories or capture them at too coarse a
level (e.g., politicians shaking hands may be reduced to ``persons''). Finally,
neither BoVW nor object detection capture properties such as saturation,
brightness, perspective, or hue, features that can also carry salience in
political contexts (e.g.  low- and high-angle shots conveying power dynamics
\citealp[e.g.,][]{hayesGretaEffectVisualising2021}, or warm versus cool color
grading shifting emotional resonance, \citealp{chenVisualFramingScience2022a,
qianConvergenceDivergenceCrossplatform2024a, hayesVisualPoliticsProtest2025}).

This said, the challenges of semantic complexity in visual data can be addressed
via an alternative path involving continuous image embeddings. Contrary to
discrete features, continuous embeddings can detect discrete elements while
simultaneously capturing nuanced relationships between them, where numerical
changes may reflect shifts in appearance, meaning, and context. For image
clustering, the task most relevant to topic modeling, researchers have developed
two prominent approaches: end-to-end models, in which a neural network
simultaneously learns a continuous representation and cluster assignments
\citep[e.g.]{xieUnsupervisedDeepEmbedding2016,
tianDeepClusterGeneralClustering2017, jiangVariationalDeepEmbedding2017a}, and
two-stage approaches, which use a pretrained vision model to first extract image
embeddings, which then serve as input to a clustering algorithm
\citep{guerinCNNFeaturesAre2017, zhangImageClusteringUnsupervised2022}. On an
intuitive level, the embedding vector can be thought of as a learned
representation that captures the hierarchical features detected by the
network---from low-level patterns to high-level semantic concepts---distilled
into a compact form, optimized for the model's training objective.  

Continuous image embeddings (pretrained or not) capture semantic complexity well
beyond the basic and literal semantic layers captured by BoVW and object
detection. Still, it is important to note that image embeddings may favor
different aspects of an image. They encode assumptions about what is relevant in
visual content: the architecture (e.g. CNN or ViT), training data and its size
(e.g. ImageNet or OpenImages), and the learning task (e.g. classification or
object detection) define the kind of information and semantic layers that will
be embedded, in theory and practice. As a consequence, the embedding model
encodes what \emph{type} of visual topic can emerge.

Models like DINO and CLIP both use ViTs as the underlying architecture and
represent state of the art for a wide variety of downstream computer vision
tasks, yet they illustrate different priorities due to their distinct training
approaches: DINO is self-supervised exclusively on images, while CLIP is trained
on image-text pairs. As a consequence, DINO can be expected to excel in
preserving similarities between images that ``look'' the same and encourage
topics that are visually coherent, while CLIP is likely to better capture
conceptual and abstract themes that cannot be understood without the context
provided by text. More specifically, DINO embeddings preserve both fine-grained
details and scene structure by learning features that are consistent under
different transformations (e.g. color jittering or cropping), generating
attention maps that align with human perception of object boundaries and
capturing complex relationships between objects (e.g. foreground/background)
along with abstract visual properties like textures and shapes
\citep{caronEmergingPropertiesSelfSupervised2021a,
oquabDinov2LearningRobust2023a}. CLIP, on the other hand, learns to associate
images and texts by predicting which caption belongs to a given image
\citep{radfordLearningTransferableVisual2021}. These image-caption pairs are
collected by crawling the web and thereby cover a much wider range of concepts
and contextual cues than standard labeled datasets. This allows CLIP to learn
comprehensive visual representations that go beyond typical object labels, both
in terms of number and type of categories. As a result, CLIP can associate
contents with geographic regions and religions (e.g., the Eiffel Tower to Paris,
a crucifix and church to Christianity), as well as detect emotions, human traits
like gender and age, abstract concepts like ``LGBTQ+'' and visual styles like
``anime'' \citep{gohMultimodalNeuronsArtificial2021}.

We do not advocate for any particular embedding model---this decision is best
made in light of specific research questions and data. However, we contend that
image embeddings in general, and pretrained embeddings in particular, offer
important advantages for the purposes of political analysis.
\citet{zhangImageClusteringUnsupervised2022} demonstrated for clustering
political image data what was well established for other domains: pretraining on
large image collections enables neural networks to learn features that are
transferable to other domains
\citep[e.g.][]{krizhevskyImageNetClassificationDeep2012,
sharifrazavianCNNFeaturesOfftheShelf2014, donahueDeCAFDeepConvolutional2014},
and follow clear scaling laws, such that downstream performance improves as
pretraining data and compute increases
\citep[e.g.][]{sunRevisitingUnreasonableEffectiveness2017,
chertiReproducibleScalingLaws2023}. \citet{zhangImageClusteringUnsupervised2022}
compared the BoVW, end-to-end DNNs and pretrained embeddings across three image
datasets focused on political memes, protest, and climate change, and found that
pretrained embeddings consistently yielded the most visually coherent clusters.
Not only is it easier, faster, and cheaper to use pretrained embeddings than
training from scratch, end-to-end models are limited in their ability to capture
meaningful information about images when data is scarce and complex, as is
likely to be the case in political analysis use scenarios.

In line with this, a handful social science studies are spearheading the use of
pretrained image embeddings with clustering in large-scale analysis of visual
data \citep{pengWhatMakesPoliticians2021, muiseSelectivelyLocalizedTemporal2022,
moosederSocialMediaLogics2023, joshiExaminingSimilarIdeologically2024}. A
sticking point is that clustering operates on a notion of single membership.
Each image is assigned to only one cluster, even though semantically rich
content can be expected to refer to several analytically important themes.
Consequently, although the image embeddings are powerful in capturing the
meaning of images, they are significantly less nimble and nuanced in rendering
visual \emph{topics}. To gain full analytical traction on visual data, we argue,
it is still desirable to integrate pretrained embeddings with the mixed
membership capacity that is a signature strength of topic modeling. The catch is
that classic topic models typically cannot take continuous features as input. In
the next section, we introduce a statistical model that addresses this obstacle.

\section{The Statistical Model}\label{sec:vstm}

\begin{figure}[htb!]
\centering
\includegraphics[page=1]{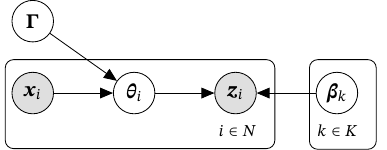}
\caption{Plate diagram of the generative model of vSTM. $\bm{\beta}_k$ is a
topic embedding, $\bm{\theta}_i$ is the topic proportions for image $i$,
$\bm{\Gamma}$ is the topic prevalence coefficients, $\bm{z}_i$ is the
image embedding and $\bm{x}_i$ are image-level covariates.}\label{fig:dgp}
\end{figure}

We posit a generative process of images, represented as image embeddings (Figure
\ref{fig:dgp}). The model is similar in spirit to the Structural Topic Model for
text in that the prevalence of topics depends on covariates through a
generalized linear model \citep{robertsStructuralTopicModels2014}. To
accommodate continuous features, we model each data point as a convex
combination of real-valued mixture components. This basic idea has been
introduced under various names as a way to cluster continuous data when
individual observations can naturally belong to multiple groups: Archetypal
Analysis \citep{cutlerArchetypalAnalysis1994}, the Partial Membership Model
\citep{hellerStatisticalModelsPartial2008}, and Mixed Membership Gaussians
\citep{giesenMixedMembershipGaussians2023}. Following topic modeling
terminology, we refer to these components (also called archetypes, prototypes,
or pure types) as \emph{topics} and their mixture weights as \emph{topic
proportions}. The \pkg{vSTM} also shares similarities with word-embedding topic
models \citep[e.g.,][]{dasGaussianLDATopic2015,
diengTopicModelingEmbedding2020b} and Contextualized Topic Models (CTMs)
extended to multimodal data \citep{bianchiPretrainingHotTopic2021,
zosaMultilingualMultimodalTopic2022}. However, it also differs in key aspects:
unlike the former \pkg{vSTM} operates on a single embedding per observation, and
unlike multimodal CTMs, our model does not require training neural networks or
having paired text-image observations. Neither approach allows directly
including covariates.

Formally, our dataset consists of $N$ images indexed by $i = 1, \ldots, N$, with
each image represented by an embedding vector $\bm{z}_i \in \mathbb{R}^D$.
Additionally, we have covariates $\bm{x}_i \in \mathbb{R}^P$ that are assumed
to influence the prevalence of visual topics. We model each image embedding
$\bm{z}_i$ as being normally distributed and generated by a mixture of $K$
topics

\begin{equation}
    \bm{z}_i \sim \text{normal}\left(\sum^K_{k=1}\theta_{i,k} \bm{\beta}_k, \bm{\Sigma}_z\right),\label{eq:likelihood}
\end{equation}

\noindent where $\bm{\theta}_i$ is a vector of \emph{topic proportions}, such
that $\theta_{i,k} \geq 0$ and $\sum_k \theta_{i,k} = 1$, and $\bm{\mathrm{B}} =
\left[\bm{\beta}_1, \dots, \bm{\beta}_K\right]^T$ is a matrix of \emph{topic
embeddings}, where $\bm{\beta}_k \in \mathbb{R}^D$. The image-specific topic
proportions thus define how much each topic contributes to $\bm{z}_i$. Note that
unlike traditional mixed-membership models, the mixing occurs at the observation
level (image), rather than at the feature level (e.g. a word in LDA). A specific
topic $\bm{\beta}_k$ is a point in the embedding space. If $\bm{z}_i$ is close
to $\bm{\beta}_k$, then the corresponding topic proportion $\theta_{i,k}$ will
also tend to be ``large''. $\bm{\Sigma}_z$ is the covariance matrix.

We sample topic proportions from a logistic-normal distribution, whose mean
depends on the covariates $\bm{x}_{i} \in \mathbb{R}^P$ and a set of topic
prevalence coefficients $\bm{\Gamma} \in {\mathbb{R}^{P \times (K-1)}}$

\begin{align}
    \bm{\theta}_{i} & \sim \text{logistic-normal}(\bm{x}_{i}
    \bm{\Gamma},
    \bm{\Sigma}_{\theta}).
\end{align}

This reflects the idea that images with similar covariates are also more likely
to have similar topic distributions. Compared to the Dirichlet distribution, the
logistic-normal is more flexible and can directly model correlations between
topics through the covariance $\bm{\Sigma}_\theta$
\citep{atchisonLogisticnormalDistributionsProperties1980, blei2006correlated}.

We must also place priors on $\bm{\mathrm{B}}, \bm{\Sigma}_z,
\bm{\Sigma}_\theta$ and $\bm{\Gamma}$. We use student-t distributions for
$\gamma_{p,k} \sim \text{student-t}(\nu_\gamma, 0, \sigma_\gamma)$ and
$\beta_{k,d} \sim \text{student-t}(\nu_\beta, 0, \sigma_{\beta})$, where $\nu$
is the degrees of freedom of the student-t, and controls the tail behavior of
the distribution.\footnote{A value somewhere between 3 and 8 is a good default
choice for a weakly informative prior and is more robust to larger values than
the normal distribution, while maintaining reasonable regularization
\citep[see][]{stan2024}.} We can mean-center each dimension of $\bm{z}_d$ and
scale $\sigma_\beta$ by the standard deviation $\text{SD}(\bm{z}_d)$, to set
weakly informative prior without changing the original scale of the embeddings.\footnote{We can also divide $\bm{z}_d$ by the standard deviation, however,
this would possibly change the geometry of embedding space. Dividing by norm
$\frac{z_{i,d}}{||\bm{z}_i||}$ is another alternative, and could be combined
with something like a Von-Mises Fisher distribution.}

Similar to Gaussian Mixture models, multiple options are possible for the
covariance $\bm{\Sigma}_z$ \citep[see][]{hellerStatisticalModelsPartial2008,
gruhlTaleTwoTypes2014a}. For example, if each topic is assumed to have its own
covariance, we may use $\bm{\Sigma}_{z_i} = \left(\sum_{k=1}^K
\theta_{i,k}\bm{\Sigma}_{\beta_k}^{-1}\right)^{-1}$, where
$\bm{\Sigma}_{\beta_k}$ could be fully parameterized, diagonal
$\bm{\Sigma}_{\beta_k} = \text{diag}(\bm{\sigma}_{\beta_k})$, or spherical
$\bm{\Sigma}_{\bm{\beta}_k} = \bm{\mathrm{I}}\sigma_{\beta_k}$.\footnote{With
topic-specific covariance matrices, the mean expression must also incorporate
precision weighting via inverse covariance matrices.} In our specific empirical
case (section \ref{sec:empirical}), we find that the much simpler
parameterization where $\bm{\Sigma}_z = \bm{\mathrm{I}}$ (which reduces the
likelihood to the sum squared errors), works well in practice. However, a more
flexible prior might be preferred in a different empirical setting.

Finally, we decompose the covariance matrix $\bm{\Sigma}_{\theta}$ into a
correlation matrix $\bm{\Omega}_{\theta}$ with ones on the diagonal and a vector
of standard deviations $\bm{\sigma}_{\theta}$, such that

\begin{equation}
\bm{\Sigma}_{\theta} = \text{diag}(\bm{\sigma}_{\theta}) \bm{\Omega}_{\theta} \text{diag}(\bm{\sigma}_{\theta}).
\end{equation}

\noindent We place a Lewandowski-Kurowicka-Joe (LKJ) prior on the correlation
matrix $\bm{\Omega}_{\theta} \sim \text{LKJ}(\eta_{\theta})$, where
$\eta_{\theta}>0$ controls the degree of correlation between the visual topics.
A value of $\eta_{\theta} = 1$ corresponds to a uniform distribution over
correlation matrices, while values greater than one concentrate more mass around
the diagonal \citep{lewandowskiGeneratingRandomCorrelation2009}.

\section{Inference}\label{sec:inference}

We use mean-field variational inference to approximate the posterior
distribution of the model parameters. Given the observed image embeddings
$\bm{Z}$ and covariates $\bm{X}$ we can write down the posterior as

\begin{equation}
p(\bm{\Theta}, \bm{\mathrm{B}}, \bm{\Gamma}, \bm{\Omega}_{\theta}, \bm{\sigma}_{\theta}
 \mid \bm{Z}, \bm{X}) =    \prod_{i=1}^{N}   \prod_{d=1}^D   \prod_{k=1}^K \prod_{p=1}^P \frac{p(z_{i,d}, \theta_{i,k}, \beta_{k,d}, \gamma_{p,k},
\Omega_{\theta_{k}}, \sigma_{\theta_k} \mid x_{i,p})}{p(z_{i,d} \mid x_{i,p})}, \label{eq:posterior}
\end{equation}

where hyperparameters are omitted for brevity and $\bm{\Sigma}_z$ is fixed as
in section \ref{sec:vstm}. The marginal likelihood in the denominator is
intractable but can be approximated by pursuing an alternative objective, which
for an individual image, is given by

\begin{equation}
    \mathcal{L}_{\text{ELBO}}(\phi) = \mathbb{E}_{q(\bm{\theta}_i, \bm{\mathrm{B}},
    \bm{\Gamma}, \bm{\Omega}_{\theta}, \bm{\sigma}_{\theta})} \left[ \log
p(\bm{z}_i, \bm{\theta}_i, \bm{\mathrm{B}}, \bm{\Gamma}, \bm{\Omega}_{\theta},
\bm{\sigma}_{\theta} \vert \bm{x}_i) - \log q_\phi(\bm{\theta}_i, \bm{\mathrm{B}},
\bm{\Gamma}, \bm{\Omega}_{\theta}, \bm{\sigma}_{\theta})
\right].\label{eq:elbo}
\end{equation}

The evidence lower bound (ELBO) is a lower bound on the (log) marginal
likelihood that can be maximized through optimization (see
\citealp{bleiVariationalInferenceReview2017} for a review and
\citealp{grimmerIntroductionBayesianInference2011} for an introduction to
political scientists). The first term is the log joint density, which can always
be evaluated, and the second term is the variational family of the approximate
posterior, parameterized by $\phi$. In the mean-field approximation, each factor
is independent, with a unique location $\lambda$ and scale $\nu$ for each
dimension

\begin{equation}
\scalebox{0.99}{$q_\phi(\bm{\theta}_i, \bm{\mathrm{B}}, \bm{\Gamma}, \bm{\Omega}_{\theta}, \bm{\sigma}_{\theta}) = q(\bm{\theta}_i\vert\bm{\lambda}_{\theta_i}, \bm{\nu}_{\theta_i})  q(\bm{\mathrm{B}}\vert\bm{\lambda}_B, \bm{\nu}_B)  q(\bm{\Gamma}\vert\bm{\lambda}_\Gamma, \bm{\nu}_\Gamma) q(\bm{\sigma}_\theta\vert\bm{\lambda}_{\sigma_\theta}, \bm{\nu}_{\sigma_\theta}) q(\bm{\Omega}_\theta\vert\bm{\lambda}_{\Omega_\theta}, \bm{\nu}_{\Omega_\theta})$}.
\end{equation}

\noindent We follow a similar approach to
\citet{kucukelbirAutomaticVariationalInference2015} for parameters with
constrained support in the original model, by applying appropriate
transformations to an unconstrained Gaussian space. Let $\bm{\eta}$ be a
parameter with constrained support (e.g. needs to sum to one or be positive), we
can then define a differentiable bijective transformation $f: \mathbb{R}^M
\rightarrow \text{supp}(\bm{\eta})$ and derive the correct density using the
change-of-variables formula

\begin{equation}
q(\bm{\eta}) = q(\bm{\zeta}) \left| \det \frac{\partial f^{-1}(\bm{\eta})}{\partial \bm{\eta}} \right| = \text{normal}(f^{-1}(\bm{\eta}) \vert \bm{\lambda}, \bm{\nu}) \left| \det J_{f^{-1}}(\bm{\eta}) \right|,
\end{equation}

where $\left|\det J_{f^{-1}}(\bm{\eta})\right|$ is the absolute determinant of
the Jacobian of the inverse transformation. Defining the variational family in
terms of transformed Gaussians lets us decouple the expectation in
(\ref{eq:elbo}) from the stochasticity in $q$ by recasting variational samples
as

\begin{equation}
    \bm{\zeta}^{(s)} = \bm{\lambda} + \bm{\nu}\odot\bm{\epsilon}^{(s)}, \quad \text{where } \bm{\epsilon}^{(s)} \sim \text{normal}(\bm{0}, \bm{\mathrm{I}}),
\end{equation}

such that $\bm{\zeta}^{(s)}$ is now a function of deterministic parameters
(that can be optimized) and a random variable, meaning the expectation is now
w.r.t. $\epsilon$ instead of $q$
\citep{kingmaAutoencodingVariationalBayes2013a}. This reparameterization trick
has important advantages compared to more traditional variational inference
methods. First, we can obtain an unbiased estimate of the ELBO by simply
averaging (\ref{eq:elbo}) across samples $\epsilon^{(s)} \sim
\text{normal}(0,1)$. This makes the model easy to extend, as the only
requirement is being able to evaluate log probabilities under samples from the
variational distribution. Moreover, this enables optimization via stochastic
gradient descent (SGD), which can significantly speed up training. The gradient
calculations required for SGD can be computed efficiently on GPUs by leveraging
automatic differentiation in libraries such as \pkg{JAX} and \pkg{PyTorch}
\citep{bradburyJAXComposableTransformations2018,
paszkeAutomaticDifferentiationPytorch2017}.

Lastly, two more techniques can be applied to further decrease the training time
and enable training of the \pkg{vSTM} when all parameters and data cannot fit in
memory. First, we employ minibatching where we subsample $B \ll N$ images at
each iteration. This requires scaling the contribution of the likelihood and
topic proportions to maintain unbiasedness:

\begin{align}
\tilde{\mathcal{L}}_{\text{ELBO}}(\phi) &=   \frac{N}{|\mathcal{B}|} \sum_{i \in \mathcal{B}} \frac{1}{S} \sum_{s=1}^{S} \log p\left(\bm{z}_i \mid \bm{\theta}_i^{(s)}, \bm{\mathrm{B}}^{(s)}\right) 
\nonumber \\ &+\frac{N}{|\mathcal{B}|} \sum_{i \in \mathcal{B}} \frac{1}{S} \sum_{s=1}^{S} \log p\left(\bm{\theta}^{(s)}_i \mid \bm{x}_i, \bm{\Gamma}^{(s)}, \bm{\Omega}_{\theta}^{(s)}, \bm{\sigma}_{\theta}^{(s)}\right) - \log q\left(\bm{\theta}^{(s)}_i \mid \bm{\lambda}_{\theta_i}, \bm{\nu}_{\theta_i}\right) \label{eq:mb_elbo} \\
& + \frac{1}{S}\sum_{s=1}^S \log p\left(\bm{\mathrm{B}}^{(s)}, \bm{\Gamma}^{(s)}, \bm{\Omega}^{(s)}_{\theta}, \bm{\sigma}^{(s)}_{\theta}\right) -  \log q\left(\bm{\mathrm{B}}^{(s)}, \bm{\Gamma}^{(s)}, \bm{\Omega}^{(s)}_{\theta}, \bm{\sigma}^{(s)}_{\theta}\right), \nonumber
\end{align}

where $S$ is the number of Monte Carl samples used to approximate the
expectation, $\mathcal{B}$ is a random subset of images, and the scaling factor
$\frac{N}{|\mathcal{B}|}$ properly weights the contribution of local parameters
$\bm{\theta}_i$\citep{hoffmanStochasticVariationalInference2013}. Note that
the gradient is ``doubly'' stochastic since we use both random minibatches and
Monte Carlo samples to approximate the ELBO. For some datasets, even maintaining
variational parameters for each $\bm{\theta}_i$ can become prohibitive, as
these scale with $O(NK)$. In such cases, we can employ amortization, where the
variational parameters for $\bm{\theta}_i$ are predicted by a neural network
$g_\omega$, that maps image embeddings and covariates to variational parameters:

\begin{equation}
[\bm{\lambda}_{\theta_i}, \bm{\nu}_{\theta_i}] = g_\omega(\bm{z}_i, \bm{x}_i)
\end{equation}

The network parameters $\omega$ now replace the local variational parameters,
significantly reducing the parameter space when $N$ is large, since the number
of parameters only depends on the desired flexibility of $g_\omega$ instead of
$N$. While this makes the model more scalable, it may suffer from additional
bias, stemming from an ``amortization gap''
\citep[e.g.,][]{margossianAmortizedVariationalInference2024}. We therefore
recommend amortization only when computational constraints make the fully local
approach infeasible.

\section{Quantities of Interest}\label{sec:qoi}

Rather than characterizing the full posterior distribution, our focus is on
point estimates derived from variational approximations:
$\bm{\hat{\Theta}}=\mathbb{E}_{q(\bm{\Theta})}[\bm{\Theta}]$,
$\bm{\mathrm{\hat{B}}}=\mathbb{E}_{q(\bm{\mathrm{B}})}[\bm{\mathrm{B}}]$,
$\bm{\hat{\Gamma}}=\mathbb{E}_{q(\bm{\Gamma})}[\bm{\Gamma}]$ and
$\bm{\hat{\Omega}}_{\theta}=\mathbb{E}_{q(\bm{\Omega}_\theta)}[\bm{\Omega}_\theta]$.
These quantities are often of primary interest in topic modeling, are easily
obtained, and enjoy frequentist consistency and asymptotic normality under the
conditions established in \citet{wangFrequentistConsistencyVariational2019}.

Since $q(\bm{\Gamma})$ and $q(\bm{\mathrm{B}})$ are Gaussian, their expected values
are simply the variational parameters for the mean, $\bm{\lambda}^*_\gamma$
and $\bm{\lambda}^*_\beta$, that maximize the ELBO. However, for parameters
such as $\Theta$, involving a non-linear transformation, $f$, the expectation is
not always available in closed form. And, due to Jensen's inequality, we cannot
simply use $f(\bm{\lambda^*})$ as a substitute. In these cases, we resort to
Monte Carlo estimation.

As is often the case in generalized linear models, the regression coefficients,
$\bm{\hat{\Gamma}}$, are difficult to interpret directly. Instead, we use
them to form predictions through the marginal distribution of topic proportions

\begin{equation}
\label{eq:qg}
    \mathbb{E}[\bm{\theta}_i \vert \bm{x}_i] =  \mathbb{E}\left[ \frac{\exp(\bm{\eta}_i)}{\sum_{k=1}^{K} \exp(\eta_{i,k})}\right],
\end{equation}

where $\bm{\eta}_i = [\bm{x}_i\bm{\hat{\Gamma}} +
\bm{\epsilon}_i, 0]$, $\bm{\epsilon}_i \sim \text{normal}(\bm{0},
\bm{\hat{\Sigma}}_\theta)$, and $\bm{x}_i$ can be set to some specific
covariate values, of interest to the researcher.

\section{Applying the Model}\label{sec:empirical}

We now showcase the \pkg{vSTM} and its properties in application. Aside from
substantiating the workings of the model, this section underlines how working
with any visual topic model differs from practices associated with traditional
text topic models to achieve the same goals of interpretation, validation, and
communication. Most obviously, visual topic models do not produce keywords that
can aid interpretation, but do facilitate visual inspection that can usefully
complement other heuristics.
 
Our example focuses on online climate communication. Evidence of echo chambers
and polarization is a repeated, but also contested, finding in the area (e.g.,
\citealp{williamsNetworkAnalysisReveals2015,
falkenbergGrowingPolarizationClimate2022a, treenDiscussionClimateChange2022,
vaneckOpposingPositionsDividing2024} and
\citealp{vaneckOpposingPositionsDividing2024} for an overview). The literature
focuses predominantly on text, despite it being well established that visual
communication plays a crucial role in reflecting and shaping public opinion
about climate change \citep{rossiYouSeeWhat2024}. To what extent do stakeholders
and opponents engage in separate visual worlds? Do diverging connections to
shared visual topics help explain the contested results?

\subsection{Data and Setup}\label{sec:data}

We focus on Twitter/X communication around United Nations Climate Change
Conferences (COP) 21-27. The data consists of $154,177$ unique images posted
with the \#COP2X hashtag by the most active users in the hashtag stream
($N=1,707$), defined by number of posts.

Each account was manually coded into one of six actor types (advocacy,
political, journalistic, private individuals, business, scientific) based on
profile description, following a coding scheme developed from
\citet{moosederSocialMediaLogics2023}. We also coded actor stance towards
climate action (for/against/unclear) based on general post content.  We assume
that visual content in stakeholder communication depends on such factors as
stance on climate action, role at the summit, and primary target audience on the
platform, as well as reactions to contextual factors and events
\citep{hopkeVisualizingParisClimate2018, stoddartInstagramArenaClimate2025,
moosederSocialMediaLogics2023, qianConvergenceDivergenceCrossplatform2024,
hayesGretaEffectVisualising2021, hayesVisualPoliticsProtest2025}.

To fit the \pkg{vSTM}, we first extracted a 768-dimensional image embedding for
each image using CLIP\footnote{The exact version of the model used is
\texttt{ViT-L/14@336px} from the Hugging Face Transformers library.}
\citep{radfordLearningTransferableVisual2021}. We selected CLIP to capture
conceptually coherent and not only visually coherent visual topics. 

We fit the model using \pkg{NumPyro} \citep{phanComposableEffectsFlexible2019}.
We use the Adam optimizer \citep{kingmaAdamMethodStochastic2014}, and a batch
size of $5,280$ to speed up convergence. Training one model, for a total of
$25,000$ iterations, took less than 20 minutes on a modern laptop. To inform the
prevalence of topics, we include as covariates the actor type interacted with
stance and COP year, as well as each main effect.

\subsection{Model Selection}

Before proceeding, we evaluate the quality of several models by varying the
number of topics and other hyperparameters. In this regard, the \pkg{vSTM} is no
different from other topic models, and combining heuristic quantitative
diagnostics with qualitative inspection of model outputs is also relevant here
to inform the decision.

\paragraph{Quantitative Diagnostics.}
We use three key metrics to inform the number of topics. First, we use 5-fold
cross-validation to compute perplexity, which is the exponential of the negative
log likelihood on $\bm{Z}_{\text{held-out}} \in
\mathbb{R}^{N_{\text{held-out}} \times D}$ divided by total number of elements
$N_{\text{held-out}} \times D$. Lower perplexity values indicate that the model
is better at predicting unseen data, meaning it has learned generalizable topics
as opposed to just memorizing the training data. Since topic proportions are
local to each image, we fit the $\bm{\Theta}_{\text{held-out}}$ for unseen
images while treating all other parameters as fixed to their posterior mean. 

In addition, we compute \emph{coherence} and \emph{exclusivity} scores, which
serve as a surrogate for \emph{semantic validity}---how internally coherent and
yet distinctive the topics are \citep{quinnHowAnalyzePolitical2010}. Note that
these metrics are conditional on the selected embedding (e.g., the coherence
score can refer to visual or conceptual coherence). As such, the metrics are
only meaningful to the extent that the representation space implied by the
embedding model is relevant (see Section \ref{sec:embeddings}), and they cannot
be used directly to compare models with different embeddings.

\begin{figure}[htb!]
\centering
\includegraphics[page=2]{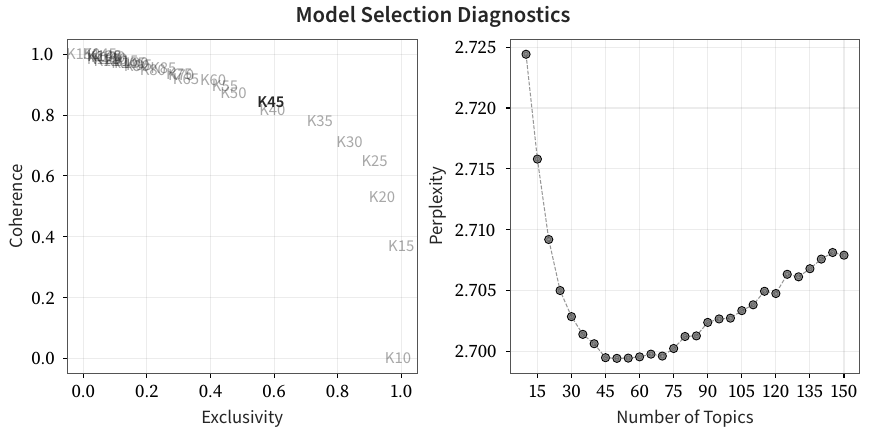}
\caption{Diagnostics for model selection. Left panel shows coherence and
exclusivity for each model, averaged over all topics. Right panel shows
perplexity, averaged over five held-out sets.}\label{fig:f1_qd}
\end{figure}

Figure \ref{fig:f1_qd} shows the diagnostics for several models, varying the
number of topics between $10$ and $150$. The left panel reports the trade-off
between coherence and exclusivity averaged over all topics (a higher value is
better). We select $K=45$ as a model that strikes a good balance between
coherence, exclusivity and perplexity. Increasing the $K$ leads to more topics
that are not meaningfully distinct; lowering it leads to less cohesive groups,
as inspecting random samples of images confirms. The perplexity curve also
suggests that $K=45$ does a good job, in relative terms, explaining unseen
data.

\paragraph{Qualitative assessment of the embedding space.} 

\begin{figure}[htb!]
\centering
\includegraphics[page=3]{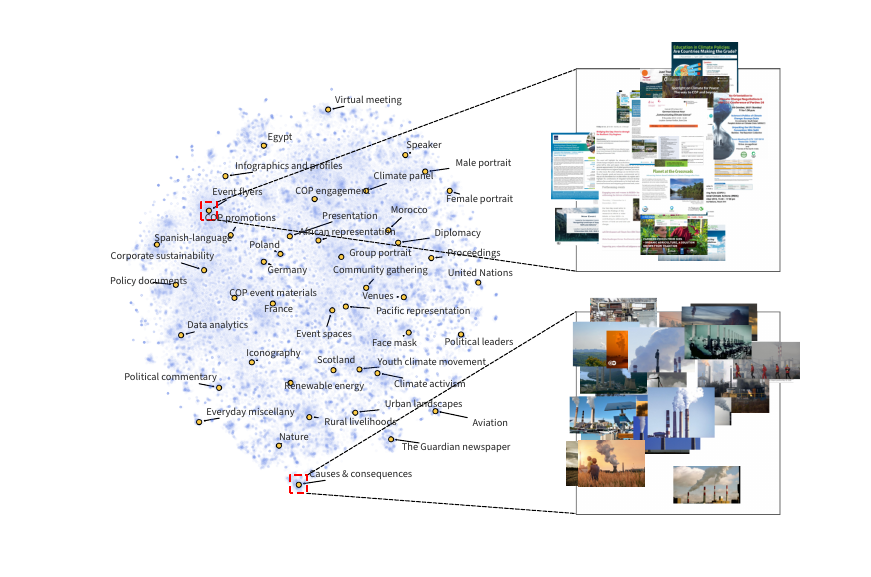}
\caption{
   Left panel: Blue points are images and yellow points topics. Right panel: 25
   randomly sampled images in the vicinity of the ``Causes/consequences'' and
   ``Event flyers'' topics.
}\label{fig:f2_tes}
\end{figure}

Qualitative evaluation of how images are distributed in the embedding space can
give substance to measures that quantify properties of interest. Figure
\ref{fig:f2_tes} shows a projection of the image and topic embeddings onto a
two-dimensional space using the t-SNE algorithm
\citep{maatenVisualizingDataUsing2008}. We display topic labels already here for
readability (see further Section \ref{sec:labels}), however, the visualization
is also useful for exploring several embedding models, before fitting the
\pkg{vSTM}.

Displaying the position of topics in relation to images enables assessing if
topics are coherent (See e.g. images close to ``Causes \& consequences''). It
also becomes clear if there are regions in the embedding space that are not
covered by topics, an indication that the number of topics may be too low. In
that case, we could sample images from those regions to assess whether they are
meaningfully represented by already identified topics. In this example, we find
that the visualization supports the diagnostics and that $K=45$ provides us with
topics that are interesting and, mostly, distinct.

\paragraph{Visualizing mixed images.} 

An additional way to probe the distribution of topics is to consider how the
model represents individual images. A strength of the mixed-membership nature of
the model is that images that do not fall neatly into a single visual topic will
not be forced to do so. The question is whether the topic distribution of mixed
images is meaningful and interpretable.

\begin{figure}[htb!]
\centering
\includegraphics[page=4]{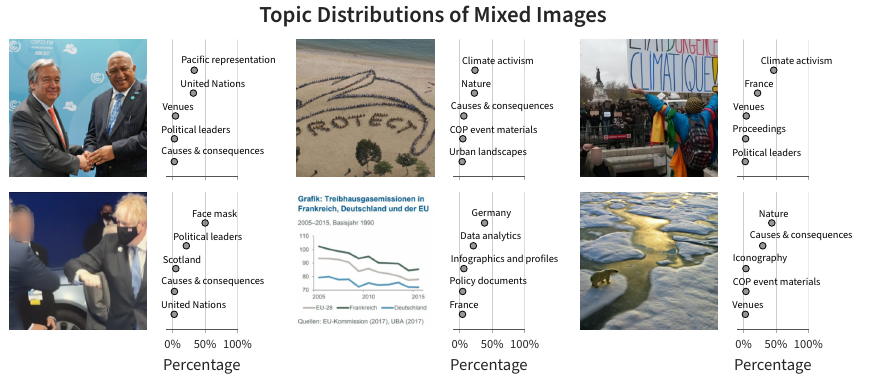}
\caption{Six images and their top five topics.}\label{fig:f3_tdmi}
\end{figure}

We analyzed images that contain at least two topics with a proportion
$\theta_{i,k} \ge 0.2$. Figure \ref{fig:f3_tdmi} showcases examples where
mixed-membership has analytical value for our purposes. Two images are
predominantly about ``Climate activism'', yet visually very different. The
second-largest topic adds additional information, enabling analysis of such
issues as tactics across time and context, or the effectiveness of different
representations of climate activism in terms of online engagement and reach.

\begin{figure}[htb!]
\centering
\includegraphics[page=5]{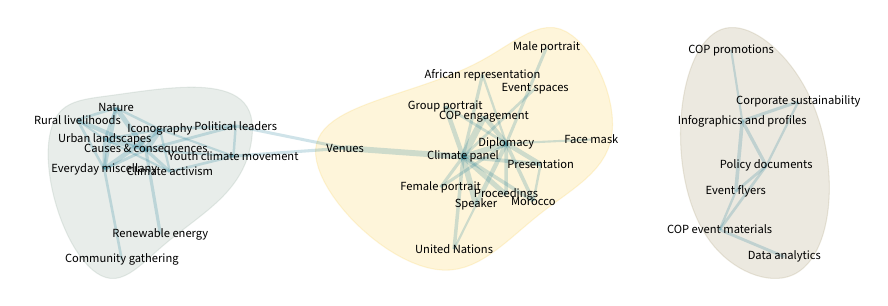}
\caption{Graph displaying a subset of positively correlated topics. Two nodes
are connected if $\bm{\Omega}_{ij} > 0.1$. The edge size is proportional to the
strength of the correlation and clusters are colored in terms of the partition
that maximizes modularity.}\label{fig:f4_tcg}
\end{figure}

Meanwhile, due to our parameterization of the covariance matrix, we can also
address how topics co-occur in images more generally, through the correlation
matrix $\bm{\Omega}_\theta$. We visualize the graph of a subset of positively
correlated topics in Figure \ref{fig:f4_tcg}, from which three meta-topics
emerge: (1) broader perspectives, e.g., climate advocacy and outdoor scenes, (2)
the conference, and (3) textual and promotional materials.

Lastly, similar to textual topic modeling, three patterns may signal a
suboptimal topic model: (1) When many images contain a high proportion of two
(or more) topics that are visually similar, suggesting that the number of topics
is too high. (2) When many images are not captured by any topic, suggesting that
the number of topics is too low. (3) Many images for which the top topics seem
accurate but the topics further down the distribution seem irrelevant. To
address this, one can consider priors that provide stronger regularization on
the topic proportions.

\subsection{Interpreting and Labeling Topics}\label{sec:labels}

\begin{center}
\begin{table}[htb!]
\caption{Summary of Topics.}
\label{tab:t1_tl}
\centering
\resizebox{\textwidth}{!}{\begin{threeparttable}[b]
\begin{tabular}{lll}
\toprule
Label & Keywords & Percentage \\
\midrule
Event flyers &  Announcement, Workshop, Virtual session, Panels & 4.94\% \\
Climate panel &  Workshop, Round table, Discussion, Slideshow & 4.89\% \\
Venues &  Plenary, Large audience, Signage, COP logo & 3.94\% \\
Policy documents &  Agreements, Reports, Commitments, Legal framework & 3.55\% \\
Causes \& consequences &  Pollution, Fossil fuels, Deforestation, Drought & 3.53\% \\
Rural livelihoods &  Agriculture, Community, Climate impact, Cultural heritage & 3.48\% \\
Data analytics &  Graphs, Statistics, Trends, Infographics & 3.25\% \\
Nature &  Landscapes, Ocean, Wildlife, Forest & 3.22\% \\
Climate activism &  Protests, Banners, Activists, Advocacy & 3.06\% \\
COP event materials &  Campaigns, Infographics, Quotes, Countdown & 2.84\% \\
Iconography &  Miscellaneous e.g. Symbols, Icons, Flags, Stock images & 2.83\% \\
Event spaces &  Exhibition, Networking zone, Conference hall, Participaiton & 2.56\% \\
Female portrait &  Close-up, Speaking, Suit, Leader & 2.56\% \\
Spanish-language &  Miscellaneous e.g. Latin America advocacy, COP25, Media & 2.53\% \\
Political leaders &  Heads of state, Diplomats, Ministers, Monarchs & 2.52\% \\
France &  Miscellaneous images e.g. French text, Media, Advocacy & 2.46\% \\
Male portrait &  Close-up, Speaking, Civil society, Leader & 2.45\% \\
Urban landscapes &  Architecture, Transportation, Landmarks, Paris & 2.40\% \\
Speaker &  Addressing, Main stage, Microphone, Close-up & 2.36\% \\
Infographics and profiles &  Quotes, Campaigns, Testimonials, Statements & 2.28\% \\
African representation &  Leaders, Advocacy, Panels, Speaker & 2.16\% \\
Diplomacy &  Negotiations, Handshake, Leaders, Meetings & 2.16\% \\
Egypt &  Miscellaneous e.g. Arabic text, Media, Profiles & 2.02\% \\
Community gathering &  Workshop, Discussions, Social Events, Group Acitivites & 1.89\% \\
Everyday miscellany &  Products, Activities, Technology, Advertisement & 1.87\% \\
Political commentary &  Cartoons, Satire, Memes, News clippings & 1.86\% \\
Group portrait &  Delegates, Awards, Networking, Smiling & 1.83\% \\
Renewable energy &  Solar panels, Wind turbines, Electric vechicles, Sustainability & 1.81\% \\
United Nations &  Guterres, Podium, UN flag, Keynote & 1.79\% \\
Morocco &  Meetings, Negotiations, Panels, People & 1.77\% \\
Corporate sustainability &  Commitments, Energy strategies, Investment, Transportation & 1.75\% \\
COP engagement &  Youth, Exhibitions, Networking, Panels & 1.64\% \\
Presentation &  Audience, Slideshow, Official, Panel & 1.50\% \\
Virtual meeting &  Zoom screens, Participant faces, Professional, Speaking & 1.48\% \\
Proceedings &  COP21, Conference hall, Panels, Official & 1.45\% \\
Scotland &  Miscellaneous e.g. Local events, Media, Advocacy & 1.40\% \\
COP promotions &  Countdowns, Initiatives, Campaigns, Participation & 1.32\% \\
Pacific representation &  Fiji, Small Island Developing States, Political leaders, Advocacy & 1.31\% \\
Germany &  Miscellaneous e.g. German text, Media, Event materials & 1.28\% \\
Youth climate movement &  Greta Thunberg, Youth protests, Protest signs, School strikes & 1.28\% \\
Face mask &  Miscellaneous e.g. Groups, Social distancing, Meetings & 1.28\% \\
The Guardian newspaper &  The Guaridan logo, Lead image, Nature, Climate impact & 1.21\% \\
Poland &  Miscellaneous e.g. Polish text, Media, Memes & 1.06\% \\
Aviation &  Map, Flight routes, Aircraft, Tracking & 0.95\% \\
\bottomrule
\end{tabular}
\begin{tablenotes}
       \item \small \textsf{Note:} The table displays the label for each topic,
       keywords representing recurring elements in representative images and the
       percentage of the topic in the data.
\end{tablenotes}
\end{threeparttable}}
\end{table}

\end{center}

We next identify keywords and labels. Unlike textual topic models, the
\pkg{vSTM} does not produce keywords that allow researchers (or readers) to
interpret, characterize, and assign labels to topics. Instead, we inspected the
top $50$ images and a random sample of $200$ images that are \emph{mostly}
associated with a topic to identify dominant concepts and features in the
images, guided by domain literature. Table \ref{tab:t1_tl} displays the topics,
keywords and the topic percentage. We show that CLIP embedding yields topics
that may be conceptually associated rather than visually similar. For example,
the topic we label as ``Aviation'' includes visually diverse images such as maps
with flight routes, flight-tracking dashboards and aircrafts. 

\subsection{Analyzing the prevalence of visual topics as a function of covariates}

Shifting into analytical application, we consider how the prevalence of visual
topics varies as a function of factors such as stance and actor type over a
number of conferences. Do stakeholders and opponents in the debate move in
separate visual worlds?

\begin{figure}[H]
\centering
\includegraphics[page=6]{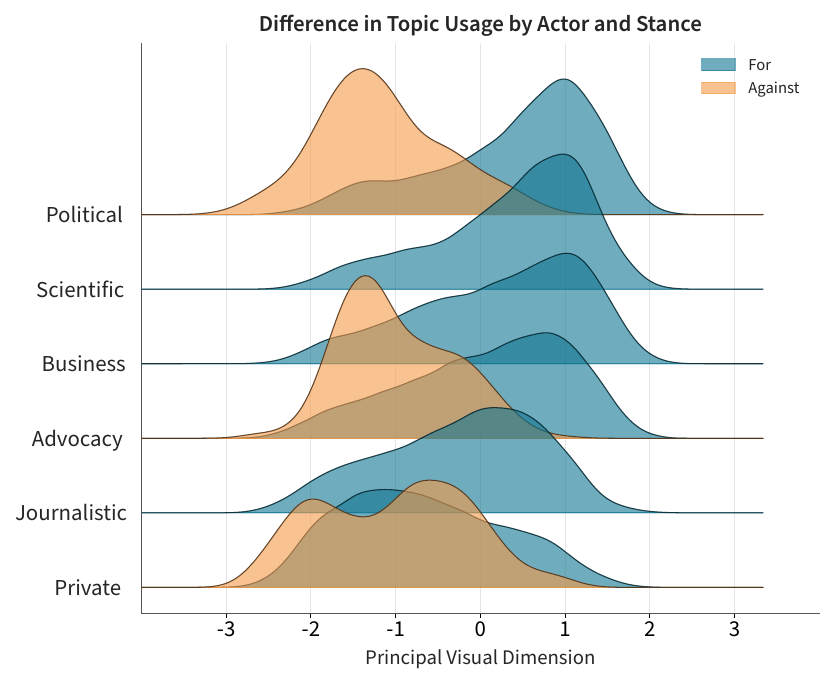}
\caption{Distribution of topic usage by actor and stance in terms of the
principal PCA component of unnormalized proportions $\bm{\lambda}_{\theta_i}$.}\label{fig:f5_sd} 
\end{figure}

We first seek an overarching view of how visual topics are wielded in our data.
We apply a principal component analysis to the image-level topic proportions,
$\bm{\theta}_i$ to identify the primary dimension of variation in topic
usage.\footnote{Principal component analysis is not well behaved under the
simplex constraint, but can be used after an additive log transformation
\citep{aitchisonPrincipalComponentAnalysis1983}. Since this is just the inverse
of the logistic-normal transform, we apply PCA to the unnormalized topic
proportions $\bm{\lambda}_{\theta_i}$.} Our quantity of interest is the
distribution of the first principal component across actor types and stances,
which we interpret as quantifying systematic differences in the frequency with
which top-producing accounts in the hashtag stream use visual topics.

Figure \ref{fig:f5_sd} suggests that types of stakeholders and opposing sides in
the debate engage with separate visual topics, but also that some visual topics
connect even as they harbor divides. The distribution for advocacy and political
actors shows two distinct peaks, suggesting that visual topics are indeed used
at different rates depending on stance for these actors. However, the overall
usage is also largely driven by the underlying composition of actors in terms of
stakeholder type. For example, ``against'' advocacy and political actors are
relatively similar to both ``for'' and ``against'' private individuals. 

\begin{figure}[htb!]
\centering
\makebox[\textwidth][c]{\includegraphics[page=7]{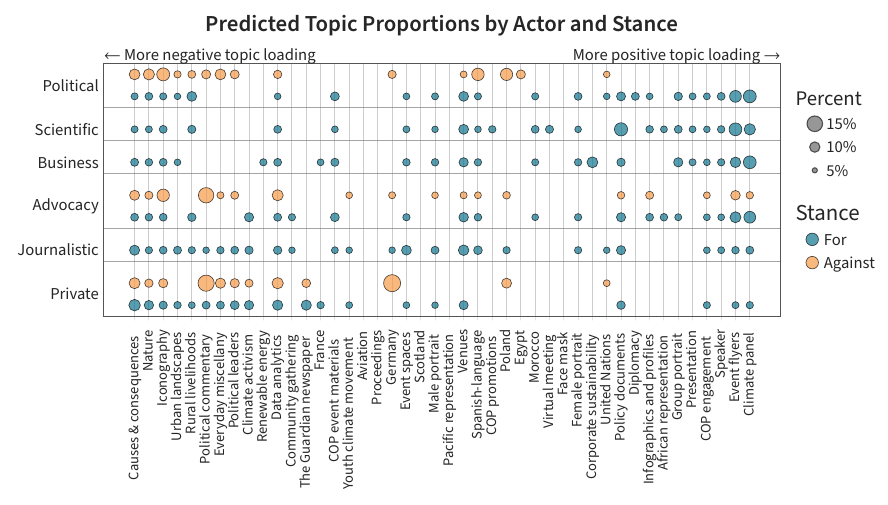}}
\caption{
    Predicted topic proportion for each topic across actor and stance. Topics
    are sorted from left-to-right, based on how they load on the principal
    visual dimension. ``Causes \& consequences'' has the most negative loading
    (left) and ``Climate panel'' the most positive. Topic proportions below
    $0.02$ are filtered out.
}\label{fig:f6_ptp}
\end{figure}

To better understand the differences, we use the PCA results to sort topics in
terms of how they load on the principal visual dimension, from most negatively
(left), to most positively (right). Figure \ref{fig:f6_ptp} shows the predicted
topic proportions (Eq. \ref{eq:qg}), by actor and stance. The size of the points
is scaled relative to the topic proportion and we prune very small proportions
for readability. Supporting the notion of separate visual worlds, the results
suggest a distance between accounts that refer to the conference (e.g., event
flyers and panels), and those with broader focus on issues and perspectives
beyond the conference (e.g., nature and political commentary).

\begin{figure}[htb!]
\centering
\makebox[\textwidth][c]{\includegraphics[page=8]{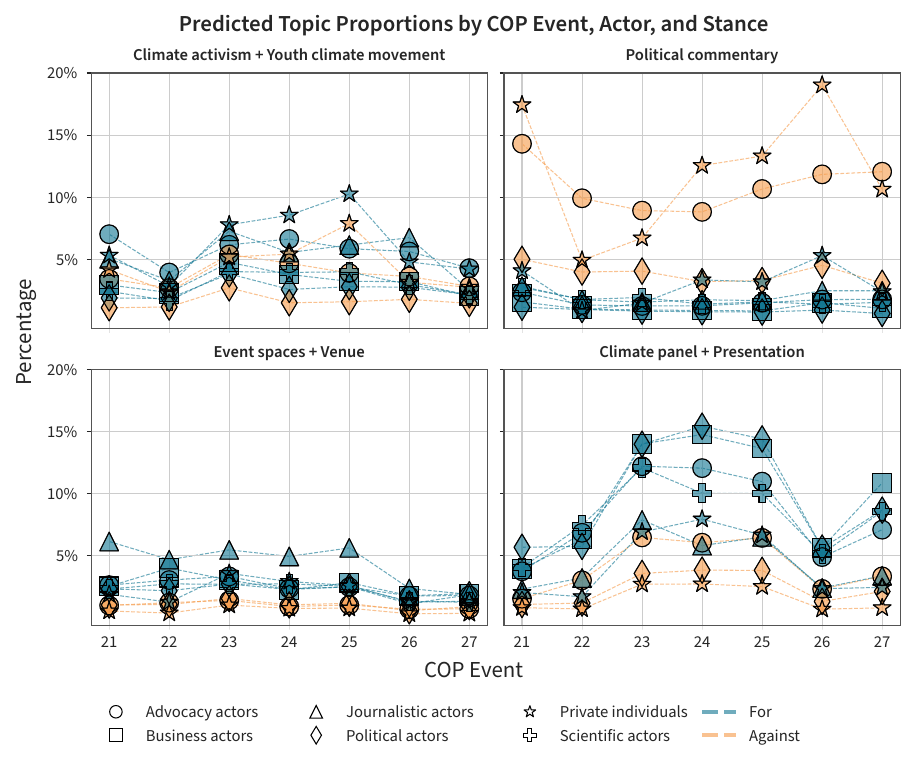}}
\caption{
    Predicted topic proportions by actor type and stance, across each COP for a
    selection of topics. Topics with a ``$+$'' are merged into one.
}\label{fig:f7_ptp}
\end{figure}

Figure \ref{fig:f7_ptp} further shows the topic proportions across actor type,
stance and COP event for a selection of topics. Top-producing private individual
accounts are likely to post images of ``climate activism'' but diverge in their
rendition. While ``for'' accounts post images of undramatic street
demonstrations, those ``against'' post images that follow a delegitimizing
``protest paradigm'' frame \citep{chanJournalisticParadigmCivil1984} with
violence (e.g. police confrontations), disruption (e.g. blocking traffic) or
spectacle (e.g. activists dressed in costumes). As a group, top-producing
politicians, business actors and scientists post a different set of topics, such
as conference scenes (e.g. ``climate panels''). Journalistic actors document
summit scenes, political elites and protests, but decreasingly with respect to
the former (e.g. ``event spaces''), correlating with the pandemic and
increasingly constrained media access \citep{hayesVisualPoliticsProtest2025}.

We note that our focus here is to show the model in application and the
empirical findings should be interpreted with caution. In this data, only a
handful of accounts (albeit highly productive ones) were found to evince a
stance “against” climate action. Recall also that the data consists of unique
images posted by highly productive accounts, not the most popular content. 

\section{Validation of Topic Coherence}

\begin{figure}[H]
\centering
\makebox[\textwidth][c]{\includegraphics[page=9]{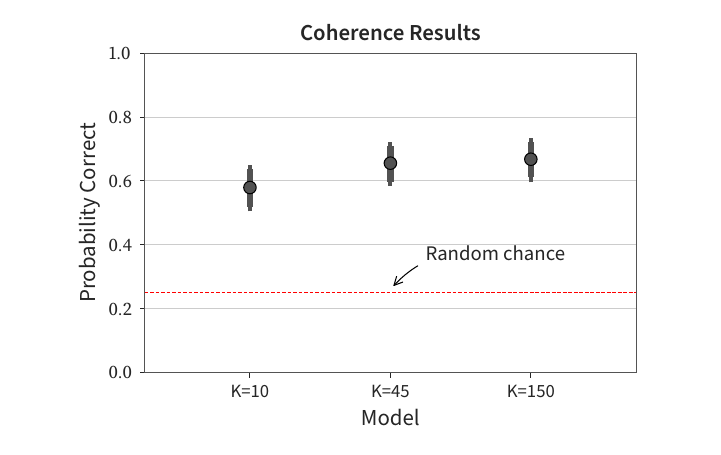}}
\caption{Predicted proportion of correct answers for the image intrusion task
across the three models. The error bars represent the $90\%$ and $95\%$ credible interval.}\label{fig:semantic_validity}
\end{figure}

We validate the coherence of topics produced by \pkg{vSTM} using experimentally
elicited human evaluations. The main results are shown in Figure
\ref{fig:semantic_validity}. 

We adapt the word intrusion task in \citet{changReadingTeaLeaves2009a,
yingTopicsConceptsMeasurement2022} to images. Specifically, we created sets of
four images where three are associated with one topic, while the fourth (the
\emph{intruder}) is associated to a different topic.  For purposes of
comparison, we also evaluate $K=10$ as an example of a model with too few topics
and $K=150$ as an example of a model with too many topics. Four researchers were
then tasked to perform $100$ evaluations per model. We fit a hierarchical
Bayesian logistic regression to predict if the intruder was correctly
identified. Similar to \citet{eshimaKeywordAssistedTopicModels2024}, we use a
common regression with fixed effects for each model. We also account for
evaluator and order effects through a random intercept. Formally, for evaluation
$i$, by evaluator $j$, on model $m$, evaluated at a randomly assigned order
index $l\in \{1,2,3\}$, the probability of identifying the intruder is given by
$P(Y_i=1)=\text{logit}^{-1}(\alpha + \psi_{m[i]} + \kappa_{j,l[i]})$, where
$\kappa$ is drawn from a common distribution $\kappa_{j,l} \sim
\text{normal}(0,\sigma_\kappa)$.

Our results show that the \pkg{vSTM} performs well across several $K$, meaning
that evaluators are able to correctly identify the intruder much more often than
what would be expected by chance. The $K=45$ and $K=150$ models perform
indistinguishably and slightly better than $K=10$. We reiterate that this
procedure quantifies a specific notion of topic coherence, not if the topics are
meaningful or scientifically interesting, so \emph{bespoke validations} remains
necessary to assess the usefulness of a visual topic model
\citep{yingTopicsConceptsMeasurement2022}. We also note that the task can become
difficult when the embedding model considers aspects in the image beyond literal
visual similarity, which might be difficult to see without domain knowledge or a
full view of all the images within a topic. How to select which images to
display to evaluators is also not trivial. For example, displaying only top
images is likely to lead to ceiling effects since, in our experience, these tend
to be very similar independent of $K$. In summary, how to best assess a visual
topic model using a similar procedure as in \citep{changReadingTeaLeaves2009a,
yingTopicsConceptsMeasurement2022} remains an open question and warrants future
research.

\section{Concluding Discussion}\label{sec:conclusion}

This paper proposes a visual topic model that captures the semantic complexity
of social and political data but retains the nuance of traditional probabilistic
topic models and the capacity to directly model the prevalence of topics as a
function of metadata. A core strength of the \pkg{vSTM} is that it can
accommodate any pretrained embedding: This greatly extends its applicability and
ease of use.

Importantly, the model's flexibility allows it to continuously improve as better
pretrained models become available. This is significant, because recent advances
in self-supervised and weakly-supervised pretraining offer a promising avenue
for \emph{off-the-shelf} image embeddings in political analysis. As exemplified
by our integration of CLIP, such embeddings display a number of properties that
can become relevant to visual topic modeling. These models also mark a paradigm
shift in computer vision toward the use of task-agnostic ``foundation models''
capable of achieving state-of-the-art performance across a wide range of
downstream tasks, on varied datasets and with little or no additional
fine-tuning (\citealp{radfordLearningTransferableVisual2021,
caronEmergingPropertiesSelfSupervised2021a}, see also
\citealp{awaisFoundationModelsDefining2025} for a review). This mirrors the
development in the text domain, where similar pretraining strategies have become
integral to the natural language processing pipeline \citep{radford2019language,
brownLanguageModelsAre2020, vaswaniAttentionAllYou2017,
devlinBERTPretrainingDeep2019a}, as increasingly recognized in political
analysis \citep[e.g.,][]{ rodmanTimelyInterventionTracking2020,
laurerLessAnnotatingMore2024, argyleOutOneMany2023}.

Relatedly, the ability to use pretrained embeddings extends the model's
potential beyond image data. The broader framework is amenable to other
embeddable modalities such as text, video, audio or social graphs. Foundational
models are commonly trained to either produce joint embedding spaces, where
references to the same object are close regardless of modality
\citep[e.g.,][]{radfordLearningTransferableVisual2021,
girdharImageBindOneEmbedding2023}, or to include a form of translation layer
that connects modality-specific encoders to large language models (LLMs),
allowing diverse inputs to be processed through a common textual interface
\citep[e.g.,][]{liBLIP2BootstrappingLanguageImage2023,
liuVisualInstructionTuning2023a}. In combination with \pkg{vSTM}, both
approaches enable truly multimodal topic modeling. The joint embedding space
approach offers a more direct path to analyzing observations of distinct (or
multiple) modalities (facilitating e.g. cross-platform research where content
formats vary), while the latter approach provides increased flexibility by
prompting for representations tailored to a specific research question
\citep{arminioLeveragingVLLMsVisual2024}.

Still, pretrained embeddings are not without limitations. One key issue is that
the \pkg{vSTM} might produce topics that reflect stereotypes and power relations
inherent to the training data of the embedding model \citep[see e.g., for
CLIP,][]{wolfeEvidenceHypodescentVisual2022,
bianchiEasilyAccessibleTexttoImage2023}. Moreover, few pretrained models are
developed with political scientists in mind, and may be tuned to produce outputs
that are normatively or commercially desirable, for its intended use, largely at
the discretion of its creator(s). This can also introduce bias, both in the
statistical and colloquial meaning of the word, especially when studying
sensitive topics. As available pretrained models increase in size and
complexity, disentangling biases introduced by the embedding (or researcher)
from those that may be of substantive interest in the data becomes increasingly
difficult. This is not unique to the \pkg{vSTM}, but points to a general need
for better understanding how pretrained models affect scientific inferences
about social and political issues. On the one hand, the researcher can choose to
be mostly agnostic about the embedding model and focus on its ability to produce
insightful and carefully validated results
\citep{grimmerMachineLearningSocial2021}. On the other hand, actively
considering the known tendencies of embedding models and how to evaluate them is
important, as this may directly inform the research design on several levels,
including ethical considerations, measurement strategies, and validation
procedures \citep{jooImageDataAutomated2022a, bergYouSeeWhat2023}.

A related issue is that image embeddings are not directly interpretable
\citep{torresFrameworkUnsupervisedSemiSupervised2024}. Although topics can be
intuitively labeled through visual inspection, a further direction to explore is
how, analogously to word lists produced by classic topic models, they can be
summarized in post-hoc representations, e.g., visual words, objects, dominant
colors or LLM descriptions. This is in principle no different from interpreting
and reporting the most frequent and exclusive (FREX) words, which are also not
direct model outputs \citep{bischof2012summarizing}. Another promising approach
is to leverage generative models to directly ``decode'' topics in the embedding
space to an interpretable representation such as an image
\citep[e.g.][]{rameshHierarchicalTextconditionalImage2022} or caption
\citep[e.g.][]{tewelZeroCapZeroShotImagetoText2022}. Recent work by
\citet{bhallaInterpretingClipSparse2024} on sparse and interpretable embeddings
is particularly interesting.

Beyond embeddings, improving the \pkg{vSTM}s capability as a measurement tool is
a critical area. For example, topic models typically struggle to reliably
capture specific concepts, often creating multiple topics of the same concept or
mixing multiple concepts in a single topic. Our model is no exception. Future
work could explore how the model can be guided toward key concepts, along the
lines of \citet{eshimaKeywordAssistedTopicModels2024} and
\citet{hurtado2024seeded}. Relatedly, variational methods trade speed for
accuracy. Particularly problematic is that the variance in the posterior
distribution is typically underestimated
\citep{wangFrequentistConsistencyVariational2019}, rendering uncertainty
quantification unreliable. While the \pkg{vSTM} is in principle a good candidate
for modern Hamiltonian Monte Carlo (e.g. the No-U-Turn Sampler
\citealp{hoffmanNoUTurnSamplerAdaptively2014}) due to its continuous parameters,
variational methods offer a practical trade-off that is often reasonable in
topic modeling contexts where researchers typically explore many models. In some
situations it is the only computationally tractable option. Nevertheless,
researchers have made important advances in reducing the errors introduced by
the variational approximation. Future iterations of the \pkg{vSTM} could benefit
from several promising directions: using more expressive variational families
\citep[e.g.,][]{ rezendeVariationalInferenceNormalizing2015}, post-hoc
corrections \citep{ vehtariParetoSmoothedImportance2024, yaoYesDidIt2018},
hybrid approaches that combine VI and MCMC
\citep[e.g.,][]{hoffmanLearningDeepLatent2017}, and alternative divergence
metrics \citep[e.g.,][]{liRenyiDivergenceVariational2016}.

\section*{Data and Code}

Code to replicate the results in the paper as well as additional supporting
information will be made available upon publication. We also intend to implement
\pkg{vSTM} as an easy-to-use open-source software package in Python.

\section*{Acknowledgments}

We thank Michelle Torres, Victoria Yantseva, Joakim Lindblad, Sophie Mainz,
Emelie Karlsson, Cristóbal Moya, Xin Shen, Diletta Goglia, Inga Wohlert, Davide
Vega, Simon Herrmann and Väinö Yrjänäinen for feedback and comments. We also
appreciated comments from participants of the 2024 workshop ``How Image-As-Data
Approaches Can Help Analysing Protest and Its Organisation: Methods and
Applications'' at Freie Universität Berlin. We thank members of the Center for
Social Media and Politics (CSMaP), New York University. The feedback from
Benjamin Guinaudeau, Jason Greenfield, Melina Munch, Chris Schwarz, Amanda
Drucker and Kylan Rutherford greatly helped improve the final draft of the
paper. This research has been funded by the Swedish Research Council for Health,
Working Life and Welfare (FORTE) grant 2021-01646 as part of the CHANSE
(Collaboration of Humanities and Social Sciences in Europe) initiative and the
Swedish Research Council grant 2021-02769. Experiments were enabled by resources
provided by the National Academic Infrastructure for Supercomputing in Sweden
(NAISS), partially funded by the Swedish Research Council through grant
agreement no. 2022-06725. MP acknowledges support from The Karl Staaff
Foundation.

\printbibliography

\end{document}